\documentclass[10pt,a4paper]{article}

\usepackage[T1]{fontenc}
\usepackage[utf8]{inputenc}
\usepackage[english]{babel}
\usepackage[margin=1.7cm,columnsep=0.6cm]{geometry}
\usepackage{amsmath,amssymb}
\usepackage{newtxtext,newtxmath}
\usepackage{microtype}
\usepackage{graphicx}
\usepackage{booktabs}
\usepackage{adjustbox}
\usepackage{titlesec}
\usepackage[numbers,sort&compress]{natbib}
\usepackage[font=small,labelfont=bf]{caption}
\usepackage{xurl}
\usepackage[hidelinks]{hyperref}

\hypersetup{
  pdftitle={MAGMA-Bench: Evaluating Long-Horizon Embodied Agents under Dynamic Tasks and Execution Failures},
  pdfauthor={Loan Bernat, Matthieu Grard, Ariane Herbulot, Florent Lamiraux},
  pdfsubject={Preprint under review},
  pdfkeywords={Benchmark, Embodied Agents, Task Planning}
}

\newcommand{\Description}[1]{}

\renewenvironment{tabular*}[2]
  {\begin{adjustbox}{max width=#1}\begin{tabular}{#2}}
  {\end{tabular}\end{adjustbox}}

\titleformat{\section}{\normalfont\large\bfseries\raggedright}{\thesection}{0.7em}{}
\titleformat{\subsection}{\normalfont\normalsize\bfseries\raggedright}{\thesubsection}{0.7em}{}
\titleformat{\paragraph}[block]{\normalfont\normalsize\bfseries}{}{0pt}{}
\titlespacing*{\section}{0pt}{2.2ex plus 0.5ex minus 0.2ex}{0.9ex}
\titlespacing*{\subsection}{0pt}{1.8ex plus 0.4ex minus 0.2ex}{0.7ex}
\titlespacing*{\paragraph}{0pt}{1.3ex plus 0.3ex minus 0.2ex}{0.4ex}

\title{\textbf{MAGMA-Bench: Evaluating Long-Horizon Embodied Agents\\
under Dynamic Tasks and Execution Failures}}
\author{
  Loan Bernat\textsuperscript{1,2}\quad
  Matthieu Grard\textsuperscript{1}\quad
  Ariane Herbulot\textsuperscript{2}\quad
  Florent Lamiraux\textsuperscript{2}\\[0.5em]
  \small\textsuperscript{1}Siléane\qquad
  \textsuperscript{2}LAAS-CNRS\\[0.3em]
  \footnotesize\texttt{l.bernat@sileane.com}
}
\date{\small\textit{Preprint --- under review}}

\begin{document}

\twocolumn[{
  \maketitle
  \begin{abstract}
    Long-horizon embodied agents must maintain task-relevant information, coordinate heterogeneous robots, and adapt their decisions as execution unfolds. Existing benchmarks investigate dynamic task execution and embodied multi-agent coordination, but provide limited evaluation of their intersection. We introduce MAGMA-Bench, an architecture-agnostic benchmark for persistent task reasoning in dynamic multi-robot environments. MAGMA-Bench evaluates agents across increasing task horizons and controlled online perturbations, including execution failures, task interruptions, and mission updates, with scenarios requiring coordination across robots with heterogeneous capabilities. It uses a structured skill-level interface to separate task-level reasoning from perception and low-level control. We evaluate locally deployable open-weight planners, multi-agent state-tracking strategies, and frontier models. Even the strongest evaluated models fail on a substantial fraction of clean episodes, and their success rates decrease further under online perturbations. Supervised adaptation greatly improves a small model, but performance remains substantially lower on longer and multi-robot missions. We release the benchmark, 261,646 task-level training examples, and an architecture-agnostic expert-trace generation pipeline.\textsuperscript{*}
  \end{abstract}
  \begin{center}
    \small\textbf{Keywords:} Benchmark, Embodied Agents, Task Planning
  \end{center}
  \begin{minipage}{\textwidth}
    \centering
    \includegraphics[width=\textwidth]{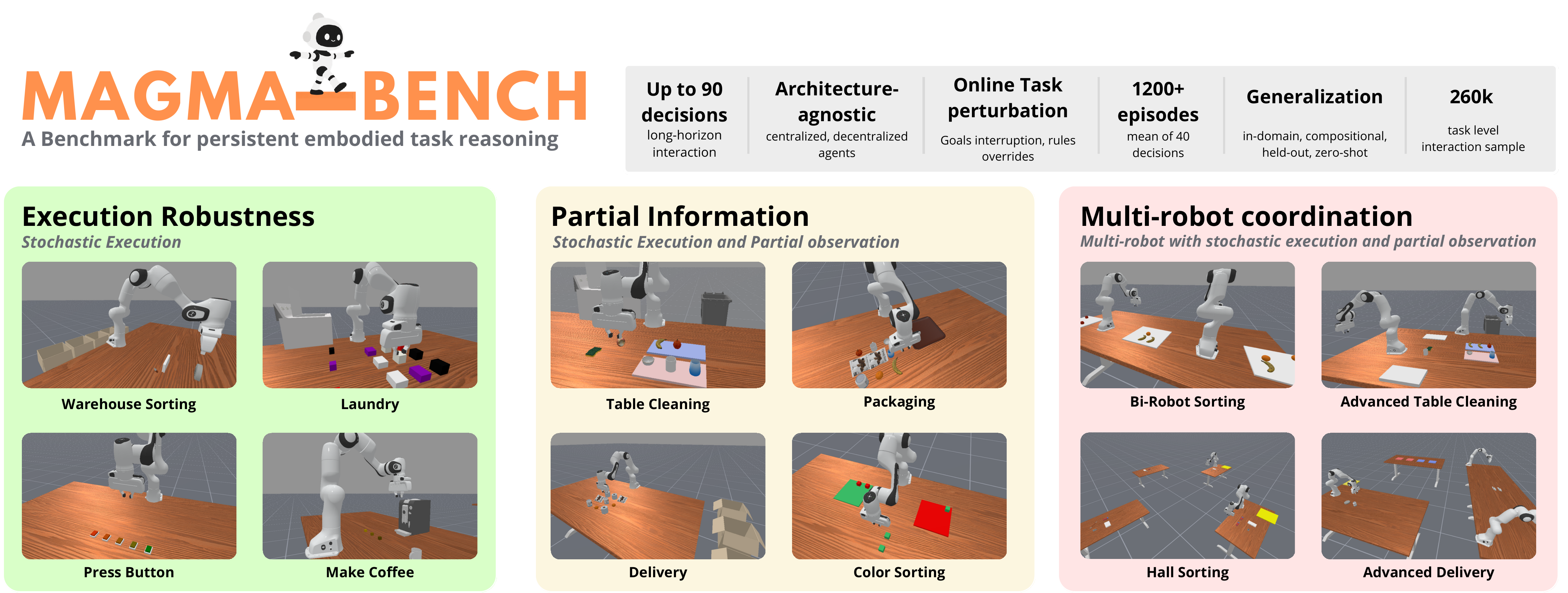}
    \captionsetup{hypcap=false}
    \captionof{figure}{The 12 scenario families in MAGMA-Bench span manipulation, navigation, and coordination with one to three robots.}
    \label{fig:magma-envs}
  \end{minipage}
  \vspace{1.5em}
}]

\begingroup
\renewcommand{\thefootnote}{\fnsymbol{footnote}}
\footnotetext[1]{\url{https://magma-rob.github.io/magma-bench}}
\endgroup

\section{Introduction}
\label{sec:introduction}

\begin{table*}[t]
    \centering
    \caption{
        Comparison with related agentic and embodied benchmarks.
        ``Abstracted'' indicates that perception and low-level control are
        separated from the evaluated task-level reasoner. Online task changes
        refer to revisions of goals or rules after execution begins.
    }
    \label{tab:benchmark_comparison}
    \small
    \setlength{\tabcolsep}{7pt}

    \begin{tabular*}{\textwidth}{@{\extracolsep{\fill}}lccccc@{}}
        \toprule
        \textbf{Benchmark}
        & \textbf{Embodied}
        & \textbf{\shortstack{Heterogeneous\\embodiments}}
        & \textbf{\shortstack{Perception /\\control abstracted}}
        & \textbf{\shortstack{Online task\\changes}}
        & \textbf{\shortstack{Controlled exec.\\perturbations}} \\
        \midrule

        MemoryAgentBench~\cite{hu2026memoryagentbench} / MemoryArena~\cite{he2026memoryarena}
        & -- & -- & n/a & -- & -- \\

        Multi-Mission Tool Bench~\cite{yu2025multimission} / InterruptBench~\cite{zou2026interruptbench}
        & -- & -- & n/a & \checkmark & -- \\

        Gaia2~\cite{froger2026gaia2}
        & -- & -- & n/a & \checkmark & \checkmark \\

        \midrule

        PARTNR~\cite{chang2025partnr}
        & \checkmark & \checkmark & \checkmark & -- & -- \\

        RoboMME~\cite{dai2026robomme} / VLABench~\cite{zhang2025vlabench}
        & \checkmark & -- & -- & -- & -- \\

        RoboCerebra~\cite{han2025robocerebra}
        & \checkmark & -- & $\sim$ & -- & \checkmark \\

        \midrule

        \textbf{MAGMA-Bench}
        & \checkmark & \checkmark & \checkmark & \checkmark & \checkmark \\

        \bottomrule
    \end{tabular*}

    \vspace{2pt}
    \footnotesize
    $\sim$ denotes a setting retaining perception in the evaluated loop.
\end{table*}

Autonomous agents increasingly use language models to interpret instructions
and select actions over extended interactions. In embodied missions, a
task-level system may coordinate robots with different skills while both the
instructions and the physical world change. An initial request may be
incomplete, requiring information from an earlier interaction or clarification
from the user. Later, a grasp may fail while another robot remains active, or a
new goal may invalidate work that has not yet been completed. Every subsequent
decision depends on what remains unknown, what has already been achieved, and
which robot can act next. The agent must therefore maintain and revise a
coherent task state throughout execution.

We call this ability \emph{persistent task reasoning}:
retaining and updating goals, constraints, assignments,
and execution progress across temporally separated events.
Existing agentic benchmarks increasingly study evolving
user instructions, interruptions, and persistent
interactions~\cite{zou2026interruptbench,froger2026gaia2}.
Meanwhile, embodied benchmarks investigate long-horizon
planning, memory, and heterogeneous agent
coordination~\cite{chang2025partnr,liu2025coherent,
han2025robocerebra}.
However, systematic evaluation of persistent task
reasoning across long-horizon interactive multi-robot
missions remains limited. Furthermore, perception and
low-level control can obscure whether a failure originates
in task-level reasoning or skill execution~\cite{magma-gen}. 
Revising a mission during multi-robot execution
is not equivalent to replanning after an action failure.
For example, suspending a delivery to prioritize another
request requires the agent to preserve unfinished work,
update the affected robot's assignment, and later resume
the delivery.
An agent must therefore track evolving goals,
constraints, and execution progress across multiple
robots, preserving valid commitments while revising
those that are no longer applicable.
Evaluating this capability requires long-horizon
episodes with controlled task and execution
perturbations, rather than isolated planning or
replanning decisions.

We introduce \textbf{MAGMA-Bench}, which instantiates this setting in 12
scenario families with one to three robots. A single decision may dispatch
skills to multiple robots and their outcomes update a shared scene while other
actions remain in progress. The interface leaves the policy architecture
unspecified. The benchmark varies nominal horizon, online task and execution
perturbations, and generalization from new in-domain instances to novel
compositions and a held-out scenario family. We evaluate locally deployable
language-model planners, state-tracking strategies, and task-specific
adaptation, and compare selected models with frontier systems on a common
215-episode subset.

In summary, our \textbf{primary contribution} is an architecture-agnostic, 
skill-level benchmark for persistent task reasoning in single- and multi-robot 
missions under controlled online perturbations. We establish the benchmark's 
utility through an empirical analysis of task horizon, state tracking, supervised
adaptation, perturbation robustness, generalization, and multi-robot
coordination, including frontier-model comparisons on a common subset.
Additionally, we release a dataset of 261,646 task-level training examples 
and an expert-trace generation pipeline that can be projected into 
architecture-specific training formats, to facilitate extension and reproducibility.

\section{Related Work}
\label{sec:related-work}

\paragraph{Memory and dynamic task execution in language agents.}
Several benchmarks establish the need to act on information that persists
across decisions. ToolSandbox~\cite{lu2025toolsandbox} tests tool use with
stateful dependencies, and $\tau$-bench~\cite{yao2025taubench} evaluates
interactions with state-changing APIs under domain policies. MemoryAgentBench
and MemoryArena~\cite{hu2026memoryagentbench,he2026memoryarena} focus on
retaining and using information over extended interactions. Multi-Mission Tool
Bench, InterruptBench, and
Gaia2~\cite{yu2025multimission,zou2026interruptbench,froger2026gaia2} move
closer to our setting by introducing mission switches, user revisions, or
dynamic environments during execution. These works make persistent task state
and online adaptation measurable in digital settings. MAGMA-Bench instead asks
how these demands change when actions have physical consequences, skill
execution is stochastic, and the available capabilities depend on the selected
robot.

\paragraph{Long-horizon reasoning in embodied agents.}
VLABench~\cite{zhang2025vlabench} evaluates language-conditioned manipulation
across diverse multi-step tasks, while RoboMME~\cite{dai2026robomme} targets
temporal, spatial, object, and procedural memory in robotic policies.
RoboCerebra~\cite{han2025robocerebra} combines planning, reflection, and memory
with adaptation to unexpected execution conditions. These benchmarks retain
visual perception and motor execution in the evaluated loop.
PARTNR~\cite{chang2025partnr} is closer to task-level coordination: it exposes
high-level skills and evaluates planning under spatial, temporal, and
heterogeneous human--robot capability constraints, including task tracking and
recovery. Recent work such as CoMuRoS~\cite{borate2026comuros}
further demonstrates event-driven replanning in
heterogeneous robot teams, including responses to
execution failures and changing user intentions.
However, its systematic replanning evaluation focuses
on the correctness of plans revised after individual
events, rather than task completion across extended
interactive episodes with controlled perturbations. 
MAGMA-Bench focuses on whether ongoing work remains coherent when
goals, rules, or robot assignments are revised after execution begins, while
skills may fail and observations may be incomplete.

\paragraph{Positioning of MAGMA-Bench.}
These works establish complementary challenges:
agentic benchmarks examine evolving user intentions
and persistent interactions, while embodied benchmarks
study physical constraints, heterogeneous capabilities,
and execution recovery.
MAGMA-Bench brings these dimensions together in a
controlled, skill-level evaluation of long-horizon
multi-robot missions.
Unlike evaluations centered on fixed task specifications
or isolated replanning events, it measures whether an
agent can maintain coherent task execution as goals,
rules, and execution conditions evolve.
By separating task-level decisions from perception
and low-level control, the benchmark enables direct
comparison of reasoning architectures and models under
reproducible perturbations. We show a brief comparison in Table~\ref{tab:benchmark_comparison}.

\section{Persistent Task Reasoning}
\label{sec:problem-formulation}

Throughout the paper, an agent denotes the complete decision-making system evaluated by the benchmark, mapping observations to robot skill calls or user-facing responses. Its internal implementation is unrestricted and may consist of a single model or of multiple interacting software agents, possibly combined with symbolic planners and memory mechanisms.

\subsection{Problem Formulation}

We consider a task-level agent coordinating embodied robots
$\mathcal{B}=\{b_i\}_{i=1}^{N}$ in a shared environment. Each robot $b_i$
exposes skills $\mathcal{U}_i$, with an argument space $\mathcal{X}_{i,u}$ for
each skill $u\in\mathcal{U}_i$. At decision step $t$, the agent produces
$a^t=(a_{\mathrm{u}}^t,a_{\mathrm{m}}^t)$ in one of three mutually exclusive modes:
\begin{equation}
a^t=
\begin{cases}
(\mathtt{null},\{(b_i,u_i^t,x_i^t)\}_{i\in J^t}) & \text{(skill calls)},\\
(a_{\mathrm{u}}^t,\mathtt{null}) & \text{(user response)},\\
(\mathtt{null},\mathtt{null}) & \text{(wait)},
\end{cases}
\end{equation}
where $\varnothing\neq J^t\subseteq\{1,\ldots,N\}$,
$u_i^t\in\mathcal{U}_i$, and $x_i^t\in\mathcal{X}_{i,u_i^t}$.
Thus, one decision may dispatch skills to several robots, with at most one
call per robot, or issue a natural-language response to the user. 
The \texttt{wait} action issues no new command and advances execution until
new user or environment events become available.

The agent receives $o^t=(k^t,q^t)$. Rather than assuming a single event per
decision, we represent $q^t$ as an event batch
\begin{equation}
q^t=(e_1^t,\ldots,e_{n_t}^t),
\qquad
e_j^t\in\mathcal{I}\cup\mathcal{E},
\end{equation}
where $\mathcal{I}$ denotes user events and $\mathcal{E}$ environment events,
such as skill execution reports. Events are processed according to their
arrival time; events generated at the same simulator timestamp are grouped
into the same $q^t$. A new decision step is therefore event-driven and occurs
whenever one or more new events become available.

The observed task attributes $k^t\in\mathcal{K}$ provide structured
information such as container names, known object classes, and each robot's
status (\texttt{running} or \texttt{free}). The agent also has access to the
skill interfaces and to permanent rules $R_{\mathrm{p}}$, which specify fixed
task guidelines. 

Skills execute asynchronously and modify a shared physical state $w^t$.
Previously dispatched skills may continue executing while the agent processes
later events or responds to the user. Consequently, a new instruction or
execution report may arrive while other robots remain active, and actions
dispatched at the same decision may complete at different times. The agent
must therefore reason from an event stream rather than assuming synchronous
step-by-step execution.

\subsection{Latent Task State}

We distinguish the physical state of the environment from the persistent
semantic state of the task. Let
$s^t=(w^t,z^t)$
denote the underlying state relevant to the interaction, where $w^t$ captures
the physical environment and $z^t$ captures the task-level information that
must persist across events. We define \emph{persistent task reasoning} as the
ability to retain, revise, and apply this task-level information across
temporally separated instructions and execution events while coordinating
robots whose skills may still be running.

More specifically,
\begin{equation}
z^t=(G^t,R_{\mathrm{u}}^t),
\end{equation}
where $G^t\subseteq\mathcal{G}$ represents the current goal configuration, including active and
suspended objectives, their priorities, robot assignments, and completion status, and
$R_{\mathrm{u}}^t\subseteq\mathcal{R}_{\mathrm{u}}$ is the set of active interaction-derived rules.
User instructions may add, revise, or override these goals and rules, whereas
$R_{\mathrm{p}}$ and $\{\mathcal{U}_i\}_{i=1}^{N}$ remain fixed throughout the episode.

Neither component of $s^t$ is directly provided to the agent. Instead, the agent must infer the task-relevant information required for its
decisions from $o^t$, $R_{\mathrm{p}}$, $\{\mathcal{U}_i\}_{i=1}^{N}$, and the preceding interaction history
\begin{equation}
h^t=(o^1,a^1,\ldots,o^{t-1},a^{t-1}).
\end{equation}

As an interaction progresses, retaining the complete history incurs increasing
context and computation costs and does not guarantee that task-relevant
information remains effectively accessible. The agent therefore defines its
own internal representation of the past interaction,
\begin{equation}
\label{eq:repr}
m^t=f(h^t).
\end{equation}
The mapping $f$ is unrestricted and may retain the full history, truncate it
to a context window, summarize it, maintain an explicit task state, or
retrieve information from an external memory. The evaluated policy is
\begin{equation}
\pi_\theta\!\left(
a^t\mid m^t,o^t,\{\mathcal{U}_i\}_{i=1}^{N},R_{\mathrm{p}}
\right).
\end{equation}

\section{Persistent Task Reasoning Benchmark}
\label{sec:benchmark}

To measure persistent task reasoning in robotics, MAGMA-Bench evaluates
whether an agent can retain, revise, and apply task-relevant information across
extended and stochastic executions. The benchmark systematically varies three
complementary dimensions: \textbf{execution horizon}, from short to extreme
missions; \textbf{online perturbations}, including mission updates, execution
failures, and goal interruptions; and \textbf{generalization}, from new
in-domain instances to novel compositions and a fully held-out scenario
family.

\begin{table*}[t]
\centering
\footnotesize
\caption{
Composition of MAGMA-Bench.
Entries under horizon and generalization report numbers of unique nominal
task specifications.
ID denotes in-domain evaluation, CG compositional
generalization, and HO a fully held-out scenario family.
The final column lists individual perturbation families available within each
scenario; CE is available throughout, and CO combines applicable families.
}
\label{tab:benchmark}
\begin{tabular*}{\textwidth}{@{\extracolsep{\fill}}l c ccccc ccc l@{}}
\hline
&
&
\multicolumn{5}{c}{\textbf{Horizon}}
&
\multicolumn{3}{c}{\textbf{Generalization}}
&
\\
\cline{3-7}
\cline{8-10}
\textbf{Scenario}
& \textbf{Robots}
& \shortstack{\textbf{Short}\\10--15}
& \shortstack{\textbf{Medium}\\16--30}
& \shortstack{\textbf{Long}\\31--45}
& \shortstack{\textbf{Very long}\\46--65}
& \shortstack{\textbf{Extreme}\\66--90}
& \textbf{ID}
& \textbf{CG}
& \textbf{HO}
& \textbf{Perturbation}
\\
\hline
Warehouse Sorting    & 1 & 1 & 3 & 0 & 2 & 1 & 5 & 2 & -- & MU, SE, IN \\
Color Sorting        & 1 & 1 & 0 & 2 & 3 & 2 & 6 & 2 & -- & MU, SE, IN \\
Laundry              & 1 & 1 & 2 & 2 & 1 & 2 & 6 & 2 & -- & MU, SE, IN \\
Bi-Robot Sorting     & 2 & 2 & 1 & 3 & 1 & 3 & 8 & 2 & -- & MU, SE \\
Make Coffee          & 1 & 4 & 2 & 2 & 1 & 1 & 7 & 3 & -- & MU, SE, IN \\
Clean Table          & 1 & 0 & 2 & 1 & 2 & 3 & 8 & 0 & -- & MU, SE \\
Press Buttons        & 1 & 1 & 2 & 2 & 0 & 0 & 5 & 0 & -- & MU, SE, IN \\
Delivery             & 1 & 1 & 1 & 2 & 1 & 3 & 6 & 2 & -- & MU, SE, IN \\
Advanced Clean Table & 2 & 1 & 0 & 2 & 0 & 4 & 7 & 0 & -- & SE \\
Hall Sorting         & 2 & 0 & 2 & 2 & 1 & 3 & 8 & 0 & -- & MU, SE, IN \\
Advanced Delivery    & 3 & 0 & 2 & 3 & 3 & 2 & 8 & 2 & -- & SE, IN \\
Packaging            & 1 & 1 & 1 & 3 & 1 & 4 & -- & -- & 10 & MU, SE, IN \\
\hline
\textbf{Total}       & -- & 13 & 18 & 24 & 16 & 28 & 74 & 15 & 10 & \\
\hline
\end{tabular*}
\end{table*}

\subsection{Task-Level Evaluation Setting}

An episode combines natural-language instructions, a set of available robotic
skills, and observations returned during execution. A policy acts at the task
level by invoking skills through a common API. Skill execution and perception
may be perturbed, and their outcomes are returned to the policy through
$q^t$. At each decision, the policy either selects skill calls
$a_{\mathrm{m}}^t$, issues a textual response $a_{\mathrm{u}}^t$ to report
completion or clarify an unresolved requirement, or waits for further events.

Agents interact with the environment exclusively through exposed robotic
skills, including perception primitives and physical manipulation or
navigation actions. Each skill is grounded in the simulated environment:
action skills modify the physical scene through motion-planned robot execution
using mplib~\cite{mplib}, whereas perception skills return observations derived
from the current scene. The benchmark thereby preserves a closed-loop coupling
between decisions, physical execution, and subsequent observations. We use
ManiSkill3~\cite{maniskill} as the simulator.

Tasks include direct instructions (``Do X''), ordered combinations (``Do X,
then Y''), persistent rules (``Remember that A must be stored in B by
default''), rule updates, and underspecified requests whose missing assignment
must be inferred from an earlier rule. They also include embodied questions,
such as ``How many X are in storage?'', and interruptions (``Stop X, do Y, and continue X later'') or (``Do Y when you are done''). Complete examples are provided in the
supplementary material.

\paragraph{Answer evaluation.}
Action stages are evaluated through executable simulator-state predicates and,
when applicable, constraints over the ordered action log. Open-ended answer
stages are evaluated by a fixed gpt-oss-20b judge~\cite{gptoss20}, which receives
the original question, an episode-specific semantic rule, and the model answer. Simple acknowledgements instead require a
non-empty textual response and do not invoke the judge. The runner also rejects
empty outputs and violations of the expected action-versus-text response mode.

\subsection{Robotic Interaction Model}

This abstraction makes the task-level policy independent of a particular robot
controller while retaining characteristic sources of uncertainty from robotic
deployment. Perception is imperfect: the detection pipeline may miss objects,
yielding a partial observation. Physical skills are also stochastic: motion-planned trajectories can be perturbed to
induce failures such as unsuccessful grasps or placements. These perturbations
affect the underlying environment rather than only the textual feedback. A
failed grasp leaves the object unmoved, whereas an undetected object remains
physically present and may become observable later. Agents must therefore
reason about the consequences of physical execution across subsequent
decisions.

Most skills correspond to physical actions, e.g., \texttt{take}, \texttt{put},
\texttt{clean}, or \texttt{press}. Other skills support observation and
execution management:
\begin{itemize}
    \item \textbf{detect} and \textbf{inspect} return structured information
    about the current scene. They abstract the outputs of task-specific
    perception systems, such as segmentation models or VLMs.
    \item \textbf{cancel} stops an ongoing skill so that the agent can revise a
    robot's assignment or prioritize another objective. It is the only skill
    accepted while that robot is \texttt{running}.
\end{itemize}

\subsection{Benchmark Composition}

\paragraph{Tasks.} Table~\ref{tab:benchmark} summarizes the 12 scenario
families, which involve one to three robots. Each scenario defines a physical
environment and the skills available to each robot; these skill sets may differ across robots.
 In multi-robot scenarios, skills execute
asynchronously, so an interruption or task update may affect one robot without
invalidating the ongoing work of the others.

\paragraph{Variants.} The scenarios contain 99 nominal task specifications. We derive controlled variants from each specification, sharing its initial environment configuration and task context:
\begin{itemize}
    \item \textbf{Clean} (CE): Reference condition without online perturbations. Instructions may still require clarification.
    \item \textbf{Mission Update} (MU): Assignments, rules, or user preferences change after execution begins.
    \item \textbf{Stochastic Execution} (SE): Skills may fail and perception can be incomplete.
    \item \textbf{Interruption} (IN): Goals must be suspended, reprioritized, and resumed.
    \item \textbf{Combined} (CO): Multiple applicable perturbation families occur within the same episode.
\end{itemize}
Each specification has between two and five conditions, including CE, and each
condition is instantiated with three semantically equivalent language
variations. These variations change object, robot, and skill names or
descriptions while preserving the environment configuration, task semantics,
success criteria, and nominal skill requirements. The resulting evaluation set contains 1,254
episodes. For paired evaluation, an execution-perturbation variant shares the clean
episode's initial state, stage sequence, instructions, and semantic mapping: it
differs only through frozen, stage-scoped error specifications. This fixes the
scheduled perturbations across policies, but not their realized exposure: a
fault is triggered only if the policy reaches the targeted stage and invokes a
compatible skill.

\paragraph{Horizon.} We define nominal task horizon as the number of successful high-level skill executions
required by the nominal expert solution. This quantity measures the amount of
task-level work required to complete a task independently of how an agent
schedules it. We consider short (10--15), medium (16--30), long (31--45),
very long (46--65), and extreme (66--90) horizons. In single-robot tasks, the nominal horizon generally coincides with the number
of skill-dispatch decisions. In multi-robot tasks, however, one agent decision
may dispatch skills to several robots concurrently. Online perturbations, failures, or interruptions may increase the number of
executed skills and decisions beyond the nominal horizon. Horizon is always estimated from the clean variant.

\paragraph{Generalization regimes.} Beyond new task instances, the benchmark
distinguishes three regimes relative to the released training distribution for
fine-tuned models:
\begin{itemize}
    \item \textbf{In-domain} (ID) tasks use scenario-specific task structures represented during training, while changing the concrete objects, goals, and episode configuration.
    \item \textbf{Compositional generalization} (CG) tasks recombine task mechanisms that are individually observed during training but never instantiated in that combination within the corresponding scenario family. For example, a scenario may be evaluated with an assignment rule observed during training only in another scenario.
    \item \textbf{Held-out} (HO) evaluation uses the Packaging scenario family, which is entirely excluded from training. This separation allows us to distinguish robustness to unseen instances from transfer through recombination and transfer to an unseen task family.
\end{itemize}

\subsection{Released Dataset}

Alongside the benchmark, we release a supervised training dataset and the
off-policy procedure used to generate it. Training and evaluation are produced
by separate pipelines from distinct seeds and independently constructed task
configurations. The two sets deliberately share scenario ontologies, object and skill
vocabularies, and individual request primitives. The resulting generalization
claim therefore concerns new task instances and compositions rather than
lexical or embodiment transfer.

Each sample represents the next task-level decision conditioned on the current
interaction state and can be rendered in the format required by a particular
model or architecture. The generation procedure is architecture-agnostic and
supports centralized policies as well as multi-agent architectures. For each
task, state-valid user requests and their parameters are sampled from the
current symbolic state, and scenario-specific programs synthesize an expert
trace from the corresponding state transition. The generation procedure keeps sampling requests
until the number of decisions required to complete the task exceeds 40. Targets are therefore
programmatic rather than teacher-LLM labels. The
traces are projected into next-decision SFT rows, with two augmented variants
per source row produced through instruction paraphrasing and consistent
renaming of entities, robots, tools, and parameters. A grouped 95/5 split keeps
all decisions and augmentations derived from the same task definition,
arguments, and seed in a single partition. The final export contains 261,646
training rows and 13,574 validation rows across the 11 training scenario
families.

The released training data include interruptions and mission updates, but no
stochastic skill failures or partial observations. Robustness to execution
failures and incomplete observations is therefore evaluated without direct
supervision from recovery trajectories.

\subsection{Metrics}

We report task success, partial completion, and retention of success under perturbation. Let $\mathcal{D}$ be a set of evaluation episodes and $y_e\in\{0,1\}$ indicate whether episode $e$ succeeds.

\paragraph{Success Rate.}
Success Rate (SR) measures the fraction of episodes for which the complete task objective is satisfied:
\begin{equation}
\mathrm{SR}(\mathcal{D})
=
\frac{1}{|\mathcal{D}|}
\sum_{e\in\mathcal{D}} y_e.
\end{equation}

\paragraph{Goal Completion.}
To capture partial progress on unsuccessful episodes, we additionally report Goal Completion (GC). Let $G_e$ denote the set of required subgoals for episode $e$, and let $G_e^{+}\subseteq G_e$ denote the subset completed by the end of the episode.

Although instructions may be revealed progressively during execution, $G_e$ is fixed by the predefined episode specification and includes all required subgoals. For example, a later instruction to perform $Y$ after completing $X$ is already represented in the specification as the sequence $X \rightarrow Y$. When $X$ is interrupted to perform $Y$ and then resumed, the specification decomposes $X$ into an intermediate subgoal $X_1$, corresponding to the state required before interruption, and a remaining subgoal $X_2$, corresponding to its completion after $Y$. The required sequence is therefore $X_1 \rightarrow Y \rightarrow X_2$.

We first compute the completion ratio independently for each episode and then average these ratios:
\begin{equation}
\mathrm{GC}(\mathcal{D})
=
\frac{1}{|\mathcal{D}|} 
\sum_{e\in\mathcal{D}}
\frac{|G_e^{+}|}{|G_e|}.
\end{equation}
Thus, each episode contributes equally to GC irrespective of the number of subgoals it contains.

\paragraph{Conditional Robustness.}
To measure retention of nominal task success, we define Conditional Robustness (CR) using paired clean and perturbed episodes. Let $y_e^{\mathrm{CE}}$ denote success on the clean condition of task instance $e$, and $y_e^{p}$ success on its paired variant under perturbation type $p$. We define
\begin{equation}
\mathrm{CR}_{p}
=
\frac{
\sum_{e\in\mathcal{D}}
\mathbf{1}
\left[
y_e^{\mathrm{CE}}=1
\land
y_e^{p}=1
\right]
}{
\sum_{e\in\mathcal{D}}
\mathbf{1}
\left[
y_e^{\mathrm{CE}}=1
\right]
}.
\end{equation}
Here, $\mathcal{D}$ contains only instances for which both CE and variant $p$ are available. CR measures the empirical fraction of clean successes retained under $p$. Since the conditioning set depends on the evaluated agent, CR must be read together with SR.

\section{Benchmark Validation and Empirical Characterization}
\label{sec:result}

\begin{figure*}[t]
    \centering
    \includegraphics[width=\textwidth]{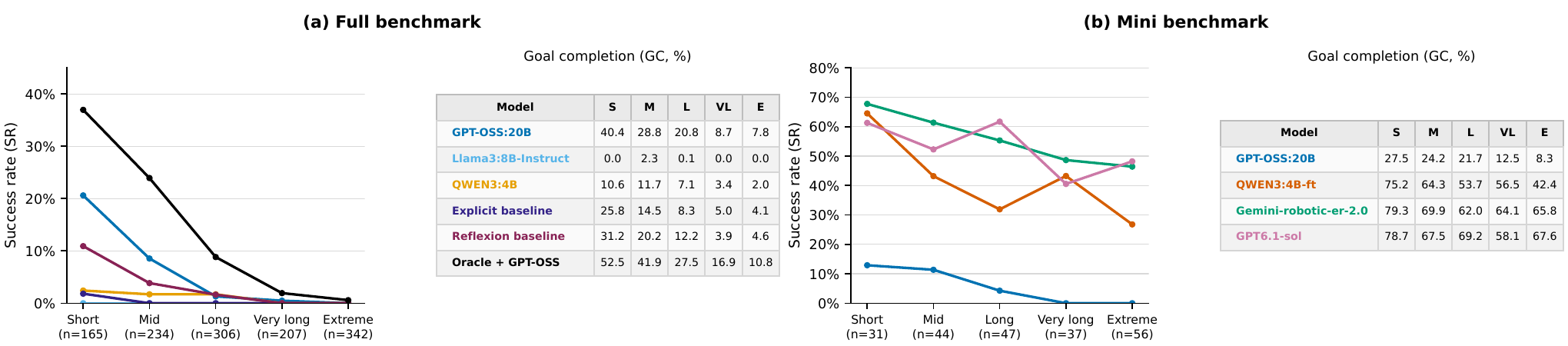}
    \caption{Success Rate over horizon aggregated over all conditions and semantic realizations. (a) Zero-shot performance on the full benchmark; (b) Frontier models and the best model on the shared 215-episode mini-benchmark.}
    \Description{Line plots compare Success Rate across five horizon buckets,
    with adjacent tables reporting Goal Completion. Zero-shot open-weight models
    approach zero success on very long and extreme tasks. On the shared
    mini-benchmark, frontier models outperform fine-tuned Qwen3-4B on longer
    missions.}
    \label{fig:horizon-analysis}
\end{figure*}

We evaluate whether MAGMA-Bench provides a useful testbed for persistent
task reasoning through four validation questions. \textbf{VQ1} asks whether
performance depends on retaining task-relevant information across temporally
separated events. \textbf{VQ2} tests whether the benchmark is learnable from
the released supervision while retaining substantial headroom.
\textbf{VQ3} examines whether its controlled difficulty dimensions, task
horizon and online perturbations, produce meaningful and model-dependent
performance changes. Finally, \textbf{VQ4} asks whether the multi-robot
scenarios expose failures associated with maintaining concurrent commitments.

\paragraph{Evaluation sets.}
Unless stated otherwise, open-weight models are evaluated on all 1,254
episodes. To limit the cost of frontier-model evaluation, we use a frozen
215-episode coverage subset. The subset approximately matches the full
benchmark's condition, horizon, and generalization-track margins, with maximum
deviations of 0.3, 2.6, and 2.2 percentage points, respectively. Every episode
is byte-identical to its counterpart in the full build. The subset contains
only one semantic realization, providing a common, cost-controlled evaluation
set rather than measuring robustness across the three semantic realizations.
We evaluate GPT-6.1-SOL~\cite{openai2026gpt61sol} and Gemini Robotics ER
2~\cite{gemini-robotics-er-2} with medium reasoning settings.

\subsection{VQ1: Can Zero-Shot Models Maintain Task State?}

Our first hypothesis is that performance depends on the agent's capacity to maintain
an accurate estimate $m^t$ of $z^t$ (Equation~\eqref{eq:repr}). We first evaluate Qwen3-4B~\cite{qwen3technicalreport},
Llama-3.1-8B-Instruct~\cite{grattafiori2024llama}, and
gpt-oss-20b~\cite{gptoss20} as single-model zero-shot policies, with $m^t$ representing the full history.
We then compare three dual-agent variants built from two zero-shot gpt-oss-20b instances: a
state manager produces a representation for an action dispatcher, whose prompt
matches the single-model policy except for instructions describing that
representation. The variants differ only in the state $m^t$ supplied by the first
model:
\begin{itemize}
    \item \textit{reflection}: a free-form paragraph produced at every step;
    \item \textit{explicit}: a structured state maintained through operations
    such as adding facts and adding or removing goals;
    \item \textit{oracle}: an oracle-provided task-state description with lists of remaining and completed goals. The oracle replaces the first model. No privileged information is leaked; only revealed goals are included.
\end{itemize}

Figure~\ref{fig:horizon-analysis}(a) shows low zero-shot success for all three
single-model baselines, even on short missions. The generated \textit{reflection} and \textit{explicit} variants
also underperform their single-model gpt-oss-20b counterpart. Qualitative trace
inspection reveals two recurring failure modes. First, conditioning the
dispatcher on an incomplete state estimate creates an information bottleneck
from which it cannot recover. Second, maintaining a structured state through
explicit update operations is itself difficult without task-specific training. By
contrast, the \textit{oracle} variant outperforms the single-model policy,
supporting the narrower conclusion that an accurate task-state representation can improve action selection relative to conditioning on the full interaction history,
but building and maintaining that state is not trivial.

\paragraph{History ablation.}
We independently restrict gpt-oss-20b to the last $K$ interaction messages.
On clean episodes, SR decreases from 5.7\% with the full available history to
3.0\% with $K=32$ and 1.0\% with $K=8$. Together, the history and oracle-state
experiments show that MAGMA-Bench decisions depend on information from
temporally distant events. This sensitivity to historical information supports the benchmark's use for evaluating persistent task reasoning.

\begin{figure*}[t]
    \centering
    \includegraphics[width=\textwidth]{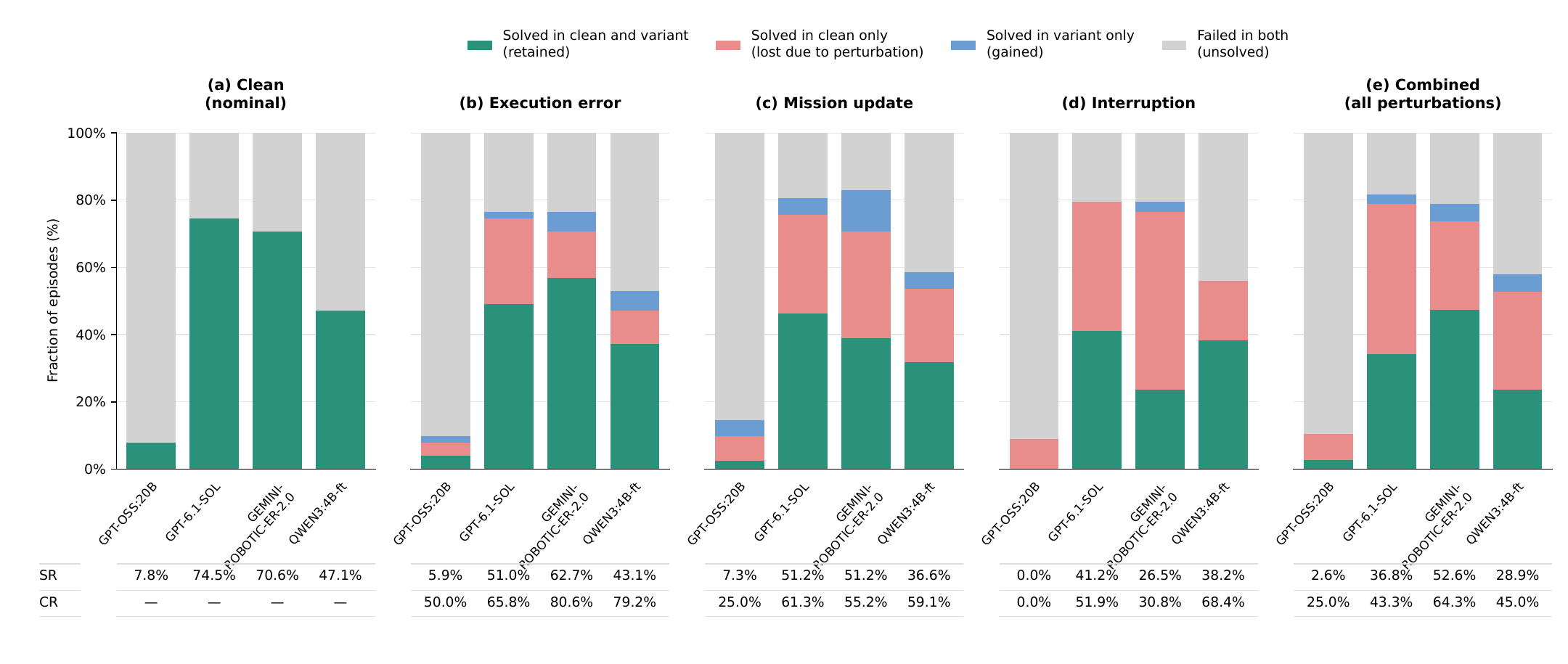}
    \caption{Paired clean and perturbed outcomes on MAGMA-Bench-mini. SR includes green and blue outcomes; CR is green divided by green plus red. Each perturbation panel uses the instances with an available clean--variant pair.}
    \Description{Stacked bars show paired episode outcomes for four models under
    clean, execution-error, mission-update, interruption, and combined conditions.
    Perturbation sensitivity differs across models, and interruptions
    sharply reduce frontier-model success.}
    \label{fig:mini-variant-analysis}
\end{figure*}

\subsection{VQ2: Is MAGMA-Bench Learnable While Retaining Headroom?}

A useful benchmark should support measurable progress without being saturated.
We therefore test whether the released supervision allows a small model to
learn the interaction protocol and task structure, and whether frontier models 
solve the benchmark.

To this end, we fine-tune Qwen3-4B with LoRA for two epochs on
the released dataset, using a learning rate of $2\times10^{-6}$, four NVIDIA RTX
6000 Ada GPUs, a batch size of two, and four gradient-accumulation steps. We
remove explicit reasoning from the targets and train the model to emit the
required textual response or tool call directly.

Table~\ref{tab:finetuning_horizon} compares the base and fine-tuned checkpoints.
Fine-tuning improves both SR and GC in every horizon bucket.
The adaptation is therefore highly effective, but the
remaining decline with horizon shows that supervision does not fully remove the
long-horizon challenge.

\begin{table}[tbp]
\centering
\caption{Qwen3-4B before and after fine-tuning on the full benchmark, by horizon.
Percentages with 95\% CIs: Wilson intervals for SR; percentile bootstrap
intervals (10,000 episode-level resamples) for GC.}
\label{tab:finetuning_horizon}
\footnotesize
\begin{tabular*}{\columnwidth}{@{\extracolsep{\fill}}lcc@{}}
\toprule
Horizon & Base & Fine-tuned \\
\midrule
\multicolumn{3}{l}{\textit{Success Rate (SR)}} \\
Short     & 2.42 [0.95, 6.07] & 72.73 [65.47, 78.95] \\
Medium    & 1.71 [0.67, 4.31] & 39.32 [33.28, 45.70] \\
Long      & 1.63 [0.70, 3.77] & 32.35 [27.36, 37.79] \\
Very Long & 0.00 [0.00, 1.82] & 29.95 [24.12, 36.51] \\
Extreme   & 0.00 [0.00, 1.11] & 21.93 [17.87, 26.61] \\
\midrule
\multicolumn{3}{l}{\textit{Goal Completion (GC)}} \\
Short     & 10.64 [7.42, 14.23] & 80.35 [75.22, 85.33] \\
Medium    & 11.74 [8.73, 15.00] & 66.58 [62.44, 70.74] \\
Long      & 7.11 [5.47, 8.95] & 52.65 [48.45, 56.95] \\
Very Long & 3.43 [2.05, 4.99] & 53.37 [48.26, 58.55] \\
Extreme   & 1.98 [1.37, 2.68] & 37.18 [33.24, 41.19] \\
\bottomrule
\end{tabular*}
\end{table}

\paragraph{Generalization.}
Qwen3-4B-FT achieves SRs of 38.19\% (95\% CI: 35.08--41.40\%) on in-domain
tasks, 33.33\% (27.14--40.16\%) on compositional tasks, and 24.00\%
(17.87--31.43\%) on the held-out Packaging family. This establishes transfer
beyond the exact training compositions and to an unseen scenario family under
the shared skill interface and ontology. Because supervised adaptation jointly
teaches interface compliance and task behavior, we do not attribute the gain
to either component in isolation.

\paragraph{Frontier Models.}
Figure~\ref{fig:horizon-analysis}(b) compares all four models on the common subset. Results for gpt-oss-20b and fine-tuned Qwen3-4B are restricted to the same episode subset. 
Gemini's SR decreases with horizon, whereas GPT's curve is non-monotonic. Fine-tuned Qwen3-4B approaches frontier-model performance on short tasks, but the gap widens on longer missions.

\subsection{VQ3: Do Horizon and Online Perturbations Expose Distinct Weaknesses?}

MAGMA-Bench varies both the amount of persistent work required by a task and
the events that modify execution online. We first use paired clean--perturbed
episodes to characterize robustness and examine horizon while
controlling for scenario composition.

\paragraph{Online perturbation effects.}
Figure~\ref{fig:mini-variant-analysis} separates absolute performance from the
retention of clean successes. The variants do not affect all models in the same way. Perturbation sensitivity differs across models. Gemini retains more clean successes under stochastic execution than under interruptions. Fine-tuned Qwen3-4B also retains a substantial fraction of its clean successes under stochastic execution, despite receiving no explicit recovery demonstrations during training. Under interruptions, Qwen achieves higher conditional retention than the frontier models, although its absolute success rate remains below GPT's and the conditioning sets differ across models.

\paragraph{Horizon effects.}
To assess horizon effects within scenarios, we compare the shortest and longest available horizon buckets using clean episodes and weighting scenarios equally. Macro-averaged goal completion decreases by 1.5--30.1 percentage points across the four models, supporting horizon length as an additional axis of task difficulty. Within-scenario trends are not uniformly monotonic; complete scenario-model-horizon results are provided in the supplementary material.

\begin{figure}[tbp]
    \centering
    \includegraphics[width=\columnwidth]{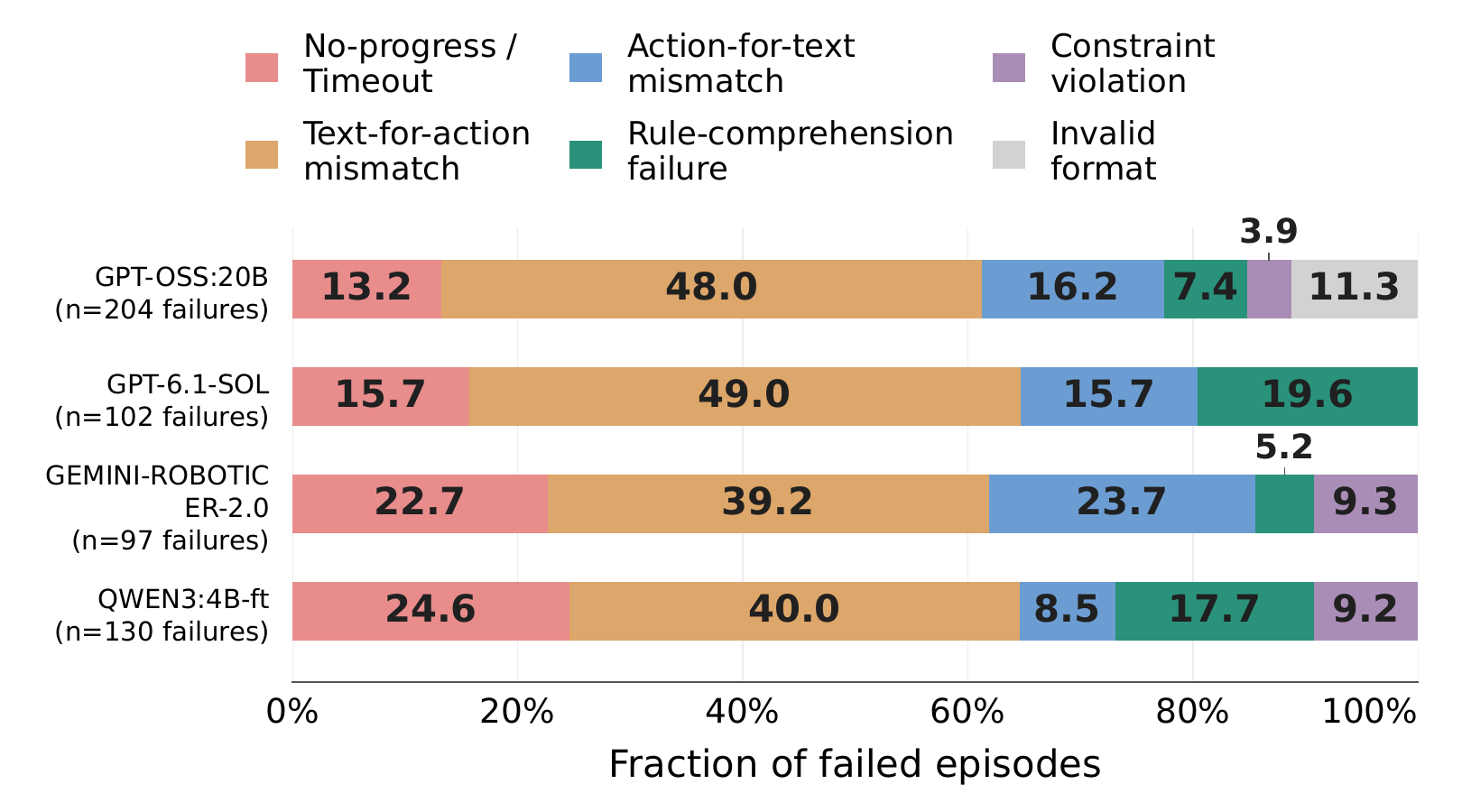}
    \caption{Terminal outcome labels among failed episodes on MAGMA-Bench-mini.
    Each bar is normalized by the corresponding model's number of failures.}
    \Description{Horizontal stacked bars partition each model's failed episodes
    into six terminal outcome categories. Text-for-action mismatches and lack of
    progress or timeouts together account for roughly 61--65\% of failures across
    all four models. The remaining categories are action-for-text mismatches,
    rule-comprehension failures, constraint violations, and invalid response formats.}
    \label{fig:terminal-failures}
\end{figure}

\paragraph{Terminal failure outcomes.}
Figure~\ref{fig:terminal-failures} groups the terminal outcomes of unsuccessful
episodes. These outcomes are assigned deterministically, and we supplement this analysis with a trace audit. Together, \emph{text-for-action} and \emph{No progress / timeout} represent 61.2--64.7\% of failures. These terminal outcomes are consistent with failures to maintain ongoing commitments: premature completion or failure to resume suspended work (an omitted subgoal), and random actions that do not contribute toward the current goal. The \emph{constraint violation} category records an action that conflicts with a task rule given earlier, while \emph{action-mismatch} represents a failure in which the model starts to act while the user is providing a rule or a partial instruction requiring clarification. Finally, 19.6\% of GPT failures and 5.2\% of Gemini failures are answers rejected by the evaluator.

\subsection{VQ4: Does MAGMA-Bench Expose Multi-Robot Coordination Failures?}
Multi-robot scenarios require a policy to maintain several concurrent
commitments while robot availability, assignments, and goals evolve
asynchronously. For this comparison, we run additional experiments with GPT-6.1-SOL to obtain comparable numbers of single- and multi-robot episodes and compare its performance with that of Qwen3-4B-FT and gpt-oss-20b. All three models are evaluated using the same linguistic realization: 34 single-robot clean episodes from the common subset and all 35 multi-robot clean specifications.

Table~\ref{tab:multi_robot} shows a pronounced model-dependent contrast. On
clean episodes, Qwen3-4B-FT decreases from 67.7\% SR on single-robot tasks to
11.4\% on multi-robot tasks, whereas GPT-6.1-SOL remains nearly unchanged
(73.5\% versus 74.3\%). GPT-OSS:20B also decreases, although its low
single-robot SR makes the contrast less informative. 
Equal weighting over the five horizon buckets leaves
a $-57.2$ point multi-minus-single gap for Qwen3-4B-FT, compared with
$+1.4$ points for GPT-6.1-SOL. Strong single-robot performance therefore
does not uniformly transfer to the benchmark's concurrent multi-robot tasks.

\begin{table}[t]
    \centering
    \caption{Clean single- versus multi-robot SR (a) and trace-based coordination
    rates (b) over evaluable multi-robot decisions or interruption obligations.
    Values are percentages; $\Delta_{\mathrm{bal}}$ equally weights horizon
    buckets and is in percentage points.}
    \label{tab:multi_robot}
    \footnotesize
    \begin{tabular*}{\columnwidth}{@{\extracolsep{\fill}}lrrr@{}}
        \toprule
        (a) Model & Single SR & Multi SR & $\Delta_{\mathrm{bal}}$ \\
        \midrule
        Qwen3-4B-FT   & 67.6 & 11.4 & $-57.2$ \\
        GPT-OSS:20B   & 11.8 &  0.0 & $-12.4$ \\
        GPT-6.1-SOL   & 73.5 & 74.3 & $+1.4$ \\
        \midrule
        (b) Model & Assign.-state err. $\downarrow$
              & Int.-reassign err. $\downarrow$
              & Resume success $\uparrow$ \\
        \midrule
        Qwen3-4B-FT & 2.04 & 81.5 & 0.0 \\
        GPT-OSS:20B & 2.90 & 30.9 & 0.0 \\
        GPT-6.1-SOL & 0.09 &  0.0 & 57.1 \\
        \bottomrule
    \end{tabular*}
\end{table}

A deterministic trace audit of multi-robot episodes, including variants, 
provides more direct evidence about the associated
coordination errors. Qwen3-4B-FT makes state-dependent assignment errors on
2.04\% of evaluable calls, compared with 0.09\% for GPT-6.1-SOL, and
incorrectly reassigns or cancels robots in 22 of 27 evaluable interruption
decisions. None of Qwen's 12 evaluable resume obligations reaches successful
completion, whereas GPT-6.1-SOL succeeds on 12 of 21.

Although the scenario populations differ, the benchmark reveals a substantial separation between policies with comparable single-robot success rates. The trace audit further identifies assignment and resumption errors within the multi-robot regime.

\section{Limitations}

MAGMA-Bench evaluates task-level decisions through structured observations and
skill calls. It does not measure visual recognition or low-level motor control,
and its results do not directly characterize end-to-end VLM or VLA policies.
The evaluated persistence concerns goals, rules, assignments, and execution
progress rather than spatial scene understanding. Also, the frontier-model
subset contains a single semantic realization and is a cost-controlled
coverage set rather than a statistical sample of the full benchmark. 

The paper focuses on presenting the benchmark and characterizing centralized
language-model policies, selected state-tracking strategies, online
perturbations, and multi-robot coordination. Although the interface also
supports symbolic planners and decentralized architectures, identifying the
best architecture for these tasks is outside the scope of this paper.

\section{Conclusion}

MAGMA-Bench evaluates whether robotic agents preserve and revise task commitments across long, asynchronous executions. The results show that task-specific supervision can substantially improve a small planner, while leaving pronounced weaknesses in multi-robot assignment and task resumption. Frontier models retain substantial headroom and exhibit different sensitivities to mission updates, interruptions, and execution failures. In particular, strong single-robot task performance does not necessarily extend to maintaining concurrent commitments under online task revisions. These findings motivate evaluating persistent task reasoning through both task outcomes and trace-level diagnostics of evolving commitments. We release the benchmark, training data, and expert-trace generation pipeline to support reproducible progress in this setting.

\clearpage
\bibliographystyle{unsrtnat}
\bibliography{ref}

\end{document}